\documentclass[10pt,journal,compsoc]{IEEEtran}
\usepackage{amsmath,amssymb,amsfonts}
\usepackage{float}
\usepackage{times}
\usepackage{algorithmic}
\usepackage{array}
\usepackage{footnote}
\usepackage{booktabs,makecell,tabularx}
\usepackage{multirow}
\usepackage{subcaption}
\usepackage{textcomp}
\usepackage{stfloats}
\usepackage{url}
\usepackage{verbatim}
\usepackage{graphicx}
\usepackage{wrapfig}
\usepackage[mathscr]{euscript}
\usepackage{setspace}
\usepackage{csquotes}
\usepackage{xcolor}

\definecolor{Gray}{gray}{0.9}
\definecolor{LightCyan}{rgb}{0.85,1,1}
\ifCLASSOPTIONcompsoc
  \usepackage[nocompress]{cite}
\else
  \usepackage{cite}
\fi
\ifCLASSINFOpdf
\else
\fi
\begin{document}

\title{Distance Weighted Trans Network for Image Completion}

\author{Pourya Shamsolmoali,
        Masoumeh Zareapoor, Huiyu Zhou, Xuelong Li, and Yue Lu
        % <-this % stops a space
\thanks{P. Shamsolmoali and Y. Lu are with the School of Electrical Engineering, East China Normal University, Shanghai, China.\protect\\
 M. Zareapoor is with the School of Automation, Shanghai Jiao Tong University, Shanghai, China.\protect\\
 H. Zhou is with the School of Computing and Mathematical Sciences, University of Leicester, Leicester, UK.\protect\\
 X. Li is with the School of Artificial Intelligence, OPtics and ElectroNics (iOPEN), and with the Key Laboratory of Intelligent Interaction and Applications, Northwestern Polytechnical University, Xi'an, China.}% <-this % stops a space
}


%\markboth{Journal of \LaTeX\ Class Files, 2023}%
%{Shell \MakeLowercase{\textit{et al.}}: Bare Advanced Demo of IEEEtran.cls for IEEE Computer Society Journals}

\IEEEtitleabstractindextext{%
\begin{abstract}
The challenge of image generation has been effectively modeled as a problem of structure priors or transformation. However, existing models have unsatisfactory performance in understanding the global input image structures because of particular inherent features (for example, local inductive prior). Recent studies have shown that self-attention is an efficient modeling technique for image completion problems. In this paper, we propose a new architecture that relies on Distance-based Weighted Transformer (DWT) to better understand the relationships between an image's components. In our model, we leverage the strengths of both Convolutional Neural Networks (CNNs) and DWT blocks to enhance the image completion process. Specifically, CNNs are used to augment the local texture information of coarse priors and DWT blocks are used to recover certain coarse textures and coherent visual structures. Unlike current approaches that generally use CNNs to create feature maps, we use the DWT to encode global dependencies and compute distance-based weighted feature maps, which substantially minimizes the problem of visual ambiguities. Meanwhile, to better produce repeated textures, we introduce Residual Fast Fourier Convolution (Res-FFC) blocks to combine the encoder's skip features with the coarse features provided by our generator. Furthermore, a simple yet effective technique is proposed to normalize the non-zero values of convolutions, and fine-tune the network layers for regularization of the gradient norms to provide an efficient training stabiliser. Extensive quantitative and qualitative experiments on three challenging datasets demonstrate the superiority of our proposed model compared to existing approaches.
\end{abstract}

}

\maketitle

\IEEEdisplaynontitleabstractindextext

\IEEEpeerreviewmaketitle

\ifCLASSOPTIONcompsoc
\IEEEraisesectionheading{\section{Introduction}\label{sec:introduction}}
\else
\section{Introduction}
\label{sec:introduction}
\fi

\IEEEPARstart{I}{mage} completion (inpainting), is a primary challenge in image processing. It involves filling of damaged or missing regions in an image with visually realistic and semantically meaningful content. It has an extensive variety of practical applications, making it a vital area of research in computer vision, including image editing, object removal, and image restoration.
In the past few years, generative neural networks have received considerable interest because of their capacity to learn complicated and high-dimensional distributions for inpainting \cite{xiang2023deep}. For such task, models like Generative Adversarial Networks (GANs) \cite{zhang2022gan} or VAEs \cite{wan2021high} have shown promising results. A GAN is made up of generator-discriminator networks set up in a zero-sum form \cite{shamsolmoali2021image}. On the other hand, VAEs \cite{phutke2023image} are a type of probabilistic models and use an encoder-decoder architecture to create a lower-dimensional representation of features from which new data samples can be made.
These models can adapt different neural networks to leverage the characteristics of the data. 
For instance, CNN's weight-sharing mechanism makes them the preferred method for various image processing tasks, but for sequential data, attention-based networks are recently the preferred designs. The attention mechanism, in particular, has recently shown solid performance on a range of tasks, including object detection \cite{carion2020end} and image reconstruction \cite{lu2022glama}.

Preliminary studies demonstrated deep learning's efficacy in image completion with semantic guidance. Conventional approaches such as \cite{yu2019free} use patch-based image matching for filling missing regions. However, deep learning-based methods with semantic guidance outperform them, producing more realistic and coherent completions. Autoencoder-based approaches are important for image completion  \cite{liu2020rethinking, yu2021diverse}. However, direct end-to-end methods give unsatisfactory results when dealing with missing parts of images. To address this issue, two-stage methods \cite{nazeri2019edgeconnect, peng2021generating} are used to learn missing structures and textures incrementally, but the outputs of these approaches have inconsistent appearances.

Indeed, the constraint of CNNs is their separate feature learning of structures and textures in images. These components interact to create the image's content, and existing methods struggle to generate visually pleasing results without understanding their coherence. Instead of relying solely on past knowledge, we adopt a neural network with a strong focus on understanding the connections between image components. This is where the attention mechanism becomes valuable, as it naturally captures many-to-many interactions and is highly effective in identifying unsupervised correlations in image distributions. By incorporating the attention mechanism, we can enhance the model's ability to create more coherent and realistic image completions.
Rather than adding attention modules to CNNs, Transformer \cite{vaswani2017attention} is another suitable framework for handling non-local modeling, in which attention is the main component of each block. To solve the issue of inpainting, some models \cite{wan2021high, zheng2022bridging} use transformer-based architectures. However, because of the complexity problem, previous studies only used transformers to infer low-resolution predictions for further processing, leading to a coarse image structure that degraded the final image quality, particularly in large missing regions.
LaMa \cite{suvorov2022resolution} is a image completion technique that uses Fast Fourier Convolution \cite{chi2020fast} to address the lack of a large receptive field. In the past, researchers faced challenges with global self-attention \cite{yu2019free, lee2021fnet} due to its computational complexity and limitations in effectively recovering repeated structures. LaMa has demonstrated better performance in this regard. However, when the missing regions become larger and extend beyond object boundaries, LaMa may encounter difficulties and produce faded structures.

This paper fills the gap in the literature by implementing a framework with an effective attention-based design. More specifically, we propose Distance-based Weighted Transformer Network (DWTNet), which considerably improves the quality of generated images.
Considering that image regions are not equally important in image inpainting tasks, the DWT calculates the weights for image tokens using the k-nearest neighbor (KNN) distancing algorithm. This can effectively reduce the visual deviations that occur in image inpainting problems. Furthermore, to minimize the intensive computation in ViTs \cite{Liu_2021_ICCV}, we adopt the concept of sparse attention \cite{child2019generating} in the DWTNet. Inaddition, to optimize the reuse of high-frequency features and improve the generation of repeating textures, we introduce a Res-FFC module. This module combines the generated coarse features and skip features from the encoder. By doing so, the Res-FFC module enhances DWTNet's ability to produce more realistic and visually coherent textures.

We designed our architecture for the task of image completion, demonstrating baseline performance on three benchmark datasets. Our key contributions are:

\begin{itemize}
\item For the recovery of coherent image structures with a tendency toward high-level interactions between pixels in an image, a new generative model with distance-based weighted feature maps is proposed.  This method is designed to fill the missing regions by taking into account all of the available contexts.
\item A module is introduced that aggregates distance-based weighted feature maps, which are more discriminative. We propose a distance-based weighted transformer to encode and calculate the weights for image tokens and improve global context inference. Moreover, to enhance the reuse of features, we built a Res-FFC unit to integrate coarse features from the generator with the encoder's skip features.
\item For better convergence and stabilizing training, a norm-regularization method is introduced by retaining the same variance of the weighted CNN gradients. This method does not require singular value decomposition to process the non-zero values of CNNs. Moreover, we performed a set of experiments on three datasets indicate that our architecture can generate high-fidelity images and outperforms current state-of-the-art inpainting approaches.
\end{itemize} 
%--------------

The rest of this paper is organised as follows: We present related studies in Section 2. In Section 3, we introduce DWTNet. The experimental results and ablation study are discussed in Section 4, and Section 5 concludes the paper.

%%%%%%%%%%%%%%%%%%%%%%%%%%%%%%%%
\section{Related Work}
\subsection{Image Completion}
When dealing with large areas of missing pixels, the majority of conventional image completion methods, such as path-based techniques \cite{yu2019free}, are unable to produce realistic images. 
In \cite{lu2022glama} and \cite{jain2023keys}, variations of LaMa \cite{suvorov2022resolution} architecture is proposed that use additional varieties of masks and a new loss functions to better capture different forms of missing information. Indeed, incorporating more damaged images during the training phase can improve the model's robustness with respect to different masks. To reduce visual distortions induced by standard convolutions, in \cite{yu2019free}, partial convolutions and gated convolutions are introduced.
Some works are also focused on the fusion of local and global knowledge. \cite{liu2020rethinking} makes use of feature equalization to integrate local and global features. 
Another successful approach is CoMod-GAN \cite{zhao2021large}, which improves the generation quality of images by adding a stochastic noise vector to the encoded representation. However, since CoMod-GAN lacks attention-related structures to expand the receptive field, the input image textures cannot be effectively reused.
These CNN-based approaches can create appropriate contents for masked areas, but they cannot guarantee the semantic content of the inpainted images is consistent.

Due to the impressive performance of the transformer design in various tasks, several transformer-based approaches have recently been developed. For instance, the first transformer-based image inpainting approach is proposed in \cite{wan2021high} to get the image prior and transfer it to a CNN. In addition, \cite{yu2021diverse} proposes a bidirectional and autoregressive transformer for incorporating the image prior. 
Although these approaches enhance performance, they have limitations due to the large-scale downsampling and generation.
In \cite{dong2022incremental, shamsolmoali2023transinpaint}, the authors develop a transformer using edge auxiliaries to acquire prior and transmit the prior with masked positional encoding to a base network. The transformer-based algorithms outperform CNN-based methods in terms of both quality and diversity. But because of their irrational structure, such as downsampling of the input image and transformer input quantization, they cause substantial information loss. On the other hand, diffusion models have recently gained significant attention due to their ability to generate high-quality images \cite{rombach2022high}. In \cite{lugmayr2022repaint}, an inpainting method based on a denoising diffusion probabilistic model is proposed. This model provides a probabilistic formulation to generate missing pixels in an image by iteratively denoising corrupted samples. To reduce the computation overhead of diffusion models, \cite{xia2023diffir} proposes an efficient image restoration model.
%%%%%%%%%%%%%
\begin{figure*}[t]
\includegraphics[width=1.92\columnwidth]{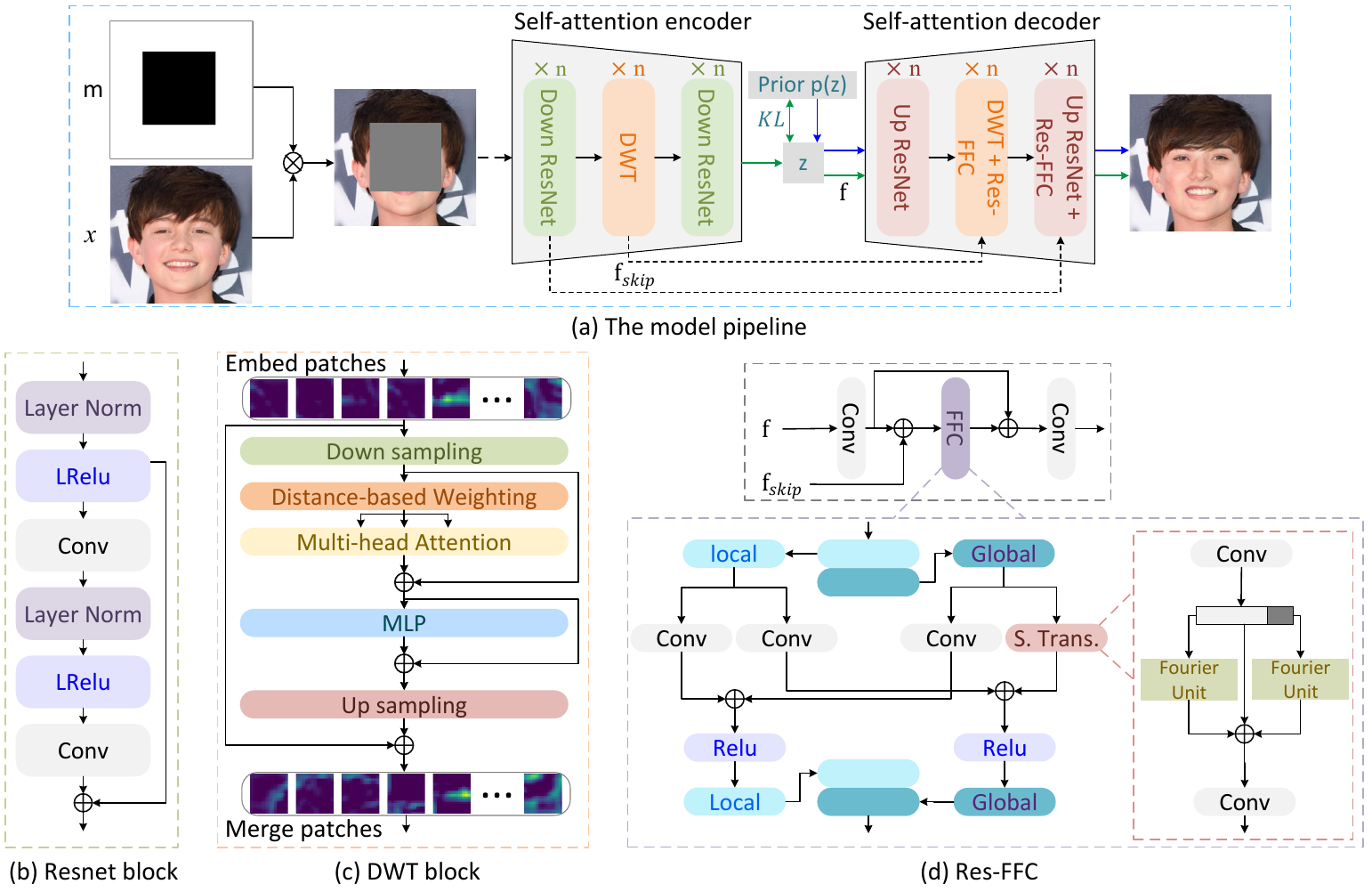}
\centering
\caption{Overview of our proposed DWTNet model. An attention-based model is proposed to parameterize the network. This biases the model toward learning correlations between input components and exploiting relationships between encoder and decoder features. Our model contains (b) Down/Up Resnet, (c) DWT, and (d) Res-FFC layers that uses Spectral Transform (S. Trans).}
\label{fig:2}
\end{figure*}
%%%%%%%%%%%%%%%%%%%%%%%%%%%%%%%%%%%%%%%%%%%%%%%%%%%%%%%%%%%%%%%%%%%

%-----------------------
\subsection{Transformers}
Transformers have been successfully used in a wide variety of vision tasks, including object detection \cite{carion2020end}, image synthesis \cite{yu2021diverse}, and image completion \cite{wan2021high}, due to their capacity to model long-range relationships. In particular, the autoregressive inference process can generate high-quality images when used for image synthetic generation \cite{wan2021high}. In TFill \cite{zheng2022bridging}, an attention-aware layer is proposed to better leverage distantly related high-frequency features, resulting in improved appearance consistency between visible and reconstructed regions. While such generation methods can produce accurate results, they involve the optimization of extra hyperparameters (such as beam size), and there is no theoretical assurance of learning the real data distribution. Choosing the appropriate level of regularization during training is more important than finding the right distribution. Beam search will only give results with sufficient diversity if the model is adequately regularized. To avoid depending on heuristics, we use an attention-based VAE to directly approximate the image distribution instead of relying on the generation process, which lacks theoretical guarantees. 

VAEs with Attention module is a new approach in machine learning and computer vision. The objective is to precisely learn the data's distribution compared to use of self-supervised methods \cite{wan2021high, skafte2019explicit}. 
Although Transformers have shown excellent performance in supervised learning tasks, their adoption in generative models has been relatively limited. However, in \cite{lin2018st}, a novel approach combines a GAN with a Transformer to effectively compose foreground objects into backgrounds, resulting in natural-looking images. In \cite{wang2019t} and \cite{shamsolmoali2023vtae}, to learn the actual data's distribution, a transformer-based latent variable technique is used in conditional VAE for the purpose of text generation. In this work, we propose a generative model with DWT that contains a bias induction toward a high-level understanding and computes the weights for image tokens to improve global context inference.
%%%%%%%%%%%%%%%%%%%%%%%%%%%
\section{Distance-based Weighted Transformer Network (DWTNet)}

In this section, we will go through the details of our DWTNet. DWTNet is a type of VAE architecture for image completion with our proposed DWT and Res-FFC layers, which serve as the primary parts of our model that parameterize the encoder and decoder. To begin, we will analyse the structure of DWTNet. This is followed by a discussion of the model, regularization, and training. In our model, the ground truth image is denoted by $x$, while the corrupted image is represented by $x_{\mathfrak m}$. 
%%%
$\mathfrak m$ is a binary matrix where $0$ denotes the missing region and $1$ denotes the observed region. This process is inherently stochastic, based on the masked image $x_{\mathfrak m}$, a conditional distribution exists $p(x\vert x_{\mathfrak m})$. By obtaining prior $z$ based on $x$ and $x_{\mathfrak m}$, then $p(x\vert x_{\mathfrak m})$ can be expressed as,

\begin{equation}
\begin{array}{lr}
p(x\vert x_{\mathfrak m})=p(x\vert x_{\mathfrak m}).p(z\vert x, x_{\mathfrak m})\\
=p(z, x\vert x_{\mathfrak m})\\
=p(z\vert x_{\mathfrak m}).p(x\vert z, x_{\mathfrak m}).
\end{array}
\label{eq:2-2}
\end{equation}
%-----------

%%%%%%%%%%%%%%%%%%%%%%%%%%%%%%%%%%%%%%%%%%%%%%%%%%%%%%%%%%%%%%%%%%%%%%%%%%

In the self-attention encoder, we extract feature vectors from the masked image $x_{\mathfrak m}$. These feature vectors are then used as input to the DWT blocks, which estimate the tokens of latent vectors for the masked regions $\mathfrak m$. To reconstruct the inpainted image, the retrieved latent vectors are given to the self-attention decoder as quantized vectors.
Our model performs sampling from the underlying distribution of appearance priors, $p(z\vert x_{\mathfrak m})$, rather than sampling from $p(x\vert x_{\mathfrak m})$. These reconstructed appearance priors provide a wider range of information for global structure and coarse textures because the DWTs are able to produce extremely high-quality representation.

As illustrated in Fig. \ref{fig:2}, our DWTNet framework consists of downsampling residual blocks, DWT blocks, upsampling residual blocks, and Res-FFC units. In the encoder network, downsampling residual blocks are used to extract tokens, and then our DWT blocks at different resolutions (with various token counts) represent long-term relationships and compute distance-based weighted feature maps. To increase the spatial resolution to the input size in the decoder layer, upsampling residual block-based reconstruction is used. To incorporate feature-to-feature context between the encoder and decoder layers, DWT blocks are used to leverage distant spatial context and effectively reduce visual deviations. Furthermore, to capture more global context information, we use the Res-FFC layers to merge generated features in the decoder with the encoder's skip features.

%%%%%%%%%%%%%%%%%%%%%%%%%%%%%
\subsection{DWTNet Architecture}
% In the next sections, we describe the components of our proposed model.

\subsubsection{Self-attention Encoder}
The downsampling Resnet blocks of the self-attention encoder take a corrupted image, and produce feature maps, which are used to generate tokens. The Resnet blocks contain convolution layers to change the input dimension and downscale the resolution. These Resnet blocks are used for two reasons. (1) To use local inductive priors for better representation and optimization in early visual processing. In addition, they supply positional information for our DWT, as demonstrated by \cite{wu2021cvt} $3\times3$ convolutions can supply adequate positional information in place of the positional embedding in ViTs. (2) The Resnet blocks are designed for quick downsampling, reducing memory usage, and computational complexity. This allows the model to efficiently process large images without a significant increase in computational requirements. In comparison to ViT's linear projection \cite{dosovitskiy2020image}, this design has been found to be more effective in handling image inpainting tasks.

\subsubsection{DWT}
As transformers are computationally intensive and lack certain inductive biases found in CNNs (e.g., translation invariance and locality \cite{dosovitskiy2020image}), we address this by using Resnet blocks to extract local features and introducing DWT as shown in Fig. \ref{fig:2} (c) to compute distance-based weighted feature maps. Given a feature map $\mathfrak{f} \in {\mathbb R}^{C\times H\times W}$, first we split the feature map into patches $\mathfrak{f}_{\textbf{p}}\in{\mathbb R}^{N \times P^2 \cdot C}$, and down-sample the patches, converting them into vectors $\textbf{x}_{\textbf{p}}\in{\mathbb R}^{N \times C}$, where $(H,W)$ denotes the feature map's resolution, $C$ represents the number of channels, $P$ denotes the down-sampling rate, and $N = HW/P^2$ shows the total number of tokens. Then, $\textbf{x}_{\textbf{p}}$ is used as the input token embeddings.
A distance-based weighted module is applied before the multi-head attention to calculate the weights for the input token embeddings $\textbf{x}_{\textbf{p}}$ by the KNN algorithm. Given a set of token embeddings $\textbf{x}_{\textbf{p}}$, the distance density $\tau_i$ of each token embedding $\textbf{x}_i$ is computed by exploiting its $k$-nearest neighbors:

%%%%%%%%%%%%%%%%%%%%%%%%%%%%%%%%%%%%
\begin{equation}
\begin{array}{lr}
    \tau_i = \text{exp} (-\frac{1}{k} \sum_{\textbf{x}_j \in \text{KNN}(\textbf{x}_i)} {\|\textbf{x}_i - \textbf{x}_j\|}^2_2),
    \label{Eq.1}
\end{array}
\end{equation}
%%%%%%%%%%%%%%%%%%%%%%%%%%%%%%%%%%%%
in which $\textbf{x}_i$ and $\textbf{x}_j$ correspond to the token embeddings, while $\text{KNN}(\textbf{x}_i)$ represents the $k$-nearest neighbors of $\textbf{x}_i$.
Consequently, we can calculate the weight for each token embedding $\textbf{x}_i$ by $w_i = \frac{1 - \tau_i}{\sum_{j = 1}^N (1 - \tau_j)}$, where $N$ is the total number of token embeddings, and $w_i$ represents the potential value of each token embedding.
%
%%%%%%%%%%%%%%%%%%%%%%%%%%%%%%%%%%%%
%\begin{equation}
%\begin{array}{lr}
 %   w_i = \frac{1 - \tau_i}{\sum_{j = 1}^N (1 - \tau_j)},
 %   \label{Eq.2}
%\end{array}
%\end{equation}
%%%%%%%%%%%%%%%%%%%%%%%%%%%%%%%%%%%%
In DWT we remove the Layer Normalization (LN) to make distancing-based weighted module performs more effectively. The DWT consists of a multi-head self-attention (MSA), and a multi-layer perceptron (MLP). For the MSA module, we use distancing-based weights in the computation of the self-attention mechanism. This can significantly minimises visual discrepancies in image completion tasks. Given the token embeddings $\textbf{x}_{\textbf{p}}$ as the input sequence of the transformer encoder, it is projected to query ($Q$), key ($K$), and value ($V$) of $h$-th attention head as follows,  

%%%%%%%%%%%%%%%%%%%%%%%%%%%%%%%%%%%%
\begin{equation}
\begin{array}{lr}
   q_h = {\textbf x_{\textbf p}} W^h_Q , ~~~ k_h = {\textbf x_{\textbf p}} W^h_K , ~~~ v_h = {\textbf x_{\textbf p}} W^h_V, 
    \label{Eq.3}
\end{array}
\end{equation}
%%%%%%%%%%%%%%%%%%%%%%%%%%%%%%%%%%%%
\begin{equation}
\begin{array}{lr}
   A_h = softmax(\frac{q_h k^T_h}{\sqrt{d_k}})
    ((\lambda_c {I}_v + {W}_c) {V}).
    \label{Eq.4}
\end{array}
\end{equation}
%%%%%%%%%%%%%%%%%%%%%%%%%%%%%%%%%%%%
{in which ${W}_{Q , K , V}$ indicate linear projection matrices for $Q$, $K$, and $V$, respectively, and the distance-based weights are applied through the diagonal matrix $W_c$ while main diagonal elements consist of each distancing-based weight $w_i$.} A scaled dot-product operation followed by softmax is used to compute the attention map ($A_h$), which determines how much frames attend to one another, while $d_k$ is the channel number of the token embeddings. $I_v$ is an identity matrix with the same rank of $v_h$. $\lambda_c$ is the scaling factor, which was set to $0.5$ in our experiments. Therefore, our DWT is formulated as follows:

\begin{equation}
\begin{array}{lr}
    \textbf{z}_{\textbf{p}}={\text MSA}(\textbf{x}_{\textbf{p}}) + \textbf{x}_{\textbf{p}}, ~~~
    \textbf{x}_{\textbf{p}} = \text{DSL}(\mathfrak{f}_{\textbf{p}}),
    \label{Eq.5}
\end{array}
\end{equation}
%%%%%%%%%%%%%%%%%%%%%%%%%%%%%%%%%%%%
\begin{equation}
\begin{array}{lr}
    \textbf{z}'_{\textbf{p}} = \text{MLP}(\textbf{z}_{\textbf{p}}) + \textbf{z}_{\textbf{p}},
    \label{Eq.6}
\end{array}
\end{equation}
%%%%%%%%%%%%%%%%%%%%%%%%%%%%%%%%%%%%
\begin{equation}
\begin{array}{lr}
    \mathfrak{f}_{\text{out}} = \text{USL}(\textbf{z}'_{\textbf{p}}) + \mathfrak{f}_{\textbf{p}}.
    \label{Eq.7}
\end{array}
\end{equation}
%%%%%%%%%%%%%%%%%%%%%%%%%%%%%%%%%%%%

DSL is average pooling to down-sample feature maps. To address the rank collapse issue, we employ an MLP consisting of two linear layers and a gaussian error linear unit activation function. Additionally, the up-sampling layer (USL) is implemented using a depth-wise separable transposed convolution.
At last, DWT reshapes the output sequence $\mathfrak{f}_{\text{out}}$ to a distancing-based weighted feature map $\mathfrak{f}_d \in {\mathbb R}^{C\times H\times W}$ to feed it to the next block.
%For the 2D feature maps, we split a feature map ${m} \in { R}^{C\times H\times W}$ into a sequence of 2D patches ${m}_p\in{ R}^{N \times P^2 \cdot C}$ and down-sample the patches into single pixels (vectors) ${x}_p\in{R}^{N \times C}$, in which $C$ is the set of channels, $(H,W)$ represent the feature map's resolution, $N = HW/P^2$ denotes the total number of patches, and $P$ indicates the down sampling rate. Then, we use ${x}_p$ as the input embeddings.
To optimize the DWT, we use a strategy similar to that used in \cite{devlin2018bert} and adopt the masked language (MaL) model. To be more precise, in the discretized input, we label the mask token indexes as $\Pi$= $\{\pi^1, \pi^2,..., \pi^J\}$, in which $J$ represents the total amount of masked tokens. $X^\Pi$ represents the number of the mask tokens in $X$, and $X^{-\Pi}$ indicates the number of the unmasked tokens. MaL aims to reduce the negative log-likelihood of $X^\Pi$ given all of the visible areas, which is written as:

%---------------
\begin{equation}
\ell_{ML}=\underset{X}{\mathbb {E}}[\frac{1}{J}\sum_{J}^{j=1}- log  ~p(x^{\pi^j}|X^{-\Pi},\theta)],
\label{eq:2-10}
\end{equation}

\noindent in which $\theta$ presents the parameters of the transformer. The incorporation of MaL with self-attention ensures that our DWT can collect all contextual information in order to estimate the probability distribution of missing regions.

\subsubsection{Self-attention Decoder}
The self-attention decoder $p_\theta(x\vert {z})$ is similar to the encoder, whose keys and queries are generated by the encoder. In experiments, we found that decoding performance improves when the encoder output is directly sent to the first decoder's layer. In our model, the autoregressive decoder $p_\theta(x\vert {z})=\prod^{\mathcal{N}}_{i=1} p_\theta(x_i\vert x_{i-1},{z})$, is used, in which $\mathcal N$ is the maximum number of sample points from a prior (such as a Binomial distribution).
In this model, two unique prior distributions are considered. $1-$The standard multivariate normal distribution with a zero mean. $2-$The multivariate normal distribution parameters use the diagonal covariance matrix to better express the prior distribution.

While autoregressive decoders make object sampling possible with different numbers of components, the prior is modeled by $p(z,t)=p(z\vert {t})p(t)$, in which $t$ is the set of sampling points. During training, we learn $p(t)$ by counting the occurrences of each sequence length, and we observe that aggregating the latent representations $z$ over all components in an image improves perceptual quality. In general, this approach is equivalent to parameterizing the posterior distribution using the encoder's output aggregated along with the dimension of the latent coordinates. In order to do this, we follow \cite{devlin2018bert}, where the first element in the decoder input is the last hidden state of the encoder for the first token, which serves as a representation of the complete sequence. 
%%%%%%%%%%%%%%%%%%%%%%%%%%%%%%%%%%%%%%%%%%%%
VAEs generally encounter a challenge called "posterior collapse" \cite{he2019lagging}. In this case, the information encoded in the latent representation is neglected by the decoder, which instead emphasizes the modes of data distribution. To address this issue and ensure that the encoder's posterior distribution accurately reflects the prior distribution, we introduce a balancing parameter $\beta$. In the training section, we will provide further details and explanations about how $\beta$ is utilized to achieve this objective.
%------------------

\subsubsection{Res-FFC}
To enhance the generation of intricate textures and meaningful semantic structures in the holes of the image, we introduce Res-FFC, as depicted in Fig. \ref{fig:2} (d), which includes a FFC layer.
The FFC layer is powered by the channel-wise Fast Fourier Transform (FFT) \cite{chi2020fast}, providing a large image-wide receptive field for more effective and efficient image completion. The FFC splits channels into two different branches:
I) Local: This branch uses general convolutions to obtain spatial information.
II) Global: The global branch utilises a Spectral Transform (S. Trans) to analyze the global structure.
The results of these branches are then combined. The Spectral Transform layer (Fig. \ref{fig:2} (d)) contains two Fourier Units (FU) to obtain both semi-global and global features. The left FU focuses on the global context, while the right FU takes one-fourth of the channels and pays more attention to the semi-global image information.
Skip connections are used in Res-FFC between encoder and decoder layers that have the same resolution scale. Res-FFC takes the features upsampled from the previous layer in the decoder (the created textures from the preceding layers) as well as the encoded skip features $\mathfrak{f}_{\text{skip}}$ (the already-existing image textures) to create the global repeating textural features. This process allows our model to use the prior coarse-level repeated textures and improve them further at the finer level. 
%%%%%%%%%%%%%%%%%%%%%%%%%%%%%%%%%%%%%%%%%%%%%%%%%%%%%%%%%%%%%%%%%%%%
\subsection{Network Regularization}

We deploy a norm-preserving convolution methodology that effectively maintains the norms of vectors using singular value regularization. Notably, this approach avoids the necessity of matrix decomposition. Specifically, in a convolution layer characterized by parameters such as input channels ($c$), output channels ($d$), and kernel size ($k$) the resulting gradient can be expressed as $\nabla_u = \hat W \nabla_v$,
%In our DWTNet, to stabilize the image completion training and accelerate the convergence, we use a norm preserving convolution that maintains norms via singular value regularization without dependency on the decomposition of the matrices. In particular, let $k$, $c$, and $d$ be the kernel size, input channels, and output channels of a convolution layer, respectively.} Therefore, the gradient can be written as $\nabla_u=\hat W \nabla_v$,
%%%%
where $u\in \mathbb R^c$ represents a vector of dimension $c$, $\nabla_u$ signifying the gradient of the input. Similarly, $v\in \mathbb R^d$ denotes a $d$-dimensional vector, while $\nabla_v$ stands for the gradient of the convolutional output. The matrix $\hat W$ is of dimensions $c\times d$ and has an important role in the backpropagation process within the convolution layer. The gradients are defined as:
%%%%%%%%%%%%%%%
%$u\in \mathbb R^c$ denotes a $c$-dimensional vector, $\nabla_u$ represents the input's gradient, $v\in \mathbb R^d$ denotes the $d$-dimensional vector, and $\nabla_v$ is the output's gradient of the convolution. $\hat W$ is an $c\times d$-dimensional matrix that performs backpropagation for the convolution layer. The gradients are defined as:
%--------

\begin{equation}
\begin{array}{lr}
\nabla_u=\sum_{i=1}^c \psi_i m_i <\nabla_v, n_i>, \\
\nabla_v=\sum_{i=1}^c n_i <\nabla_v, n_i >,
 \end{array}
  \label{eq:12}
\end{equation}
%%%%%%%%%%%
here, $\psi_i$ corresponds to a singular value of $\hat W$, while $m_i$ and $n_i$ represent the respective left and right singular vectors, and the computation of the estimated gradient norms are derived as follows:
%in which $\psi_i$ is a singular value of $\hat W$,  and $m_i$ and $n_i$ are corresponding left and right singular vectors, respectively. Therefore, the estimated norms of the gradients are calculated as:
%%%%%%%%%%%%%

\begin{equation}
\begin{array}{lr}
       \mathbb {E}[\Vert \nabla_u\Vert_2^2] =\sum_{i=1}^c \psi_i^2 \mathbb{E}[c \vert < \nabla_v, n_i > \vert^2 ],\\
       \mathbb {E}[\Vert \nabla_v\Vert_2^2] =\sum_{i=1}^d \mathbb{E}[d \vert < \nabla_v, n_i > \vert^2 ],\\
        \end{array}
        \label{eq:13}
\end{equation}
%%%%%%%%%%%%%%%%%%%%%%%%%%%%%%%%%%%%%%
To ensure uniform data dimensions, we define: $m_i\times m_j=n_i\times n_j=1$ for $i=j$, and 0 otherwise. Consequently, in order to maintain gradient norm consistency, we introduce the concept of $\mathbb{E} [\nabla_u] = \mathbb{E} [\nabla_v]$ through the allocation of non-zero values to $\psi$:
%For the equality of data dimensions, we set: $m_i\times m_j=n_i\times n_j=1$ if $i=j$, otherwise 0. Therefore, to preserve the gradients' norm, we propose to set $\mathbb{E} [\nabla_u] = \mathbb{E} [\nabla_v]$ by assigning non-zero singular values to $\psi$:

\begin{equation}
%\sigma^2 = \frac{\sum_{i=1}^d}\mu [\vert \nabla_v, n_i >\vert^2]{\sum_i \mu [\nabla_v, n_i > \vert^2]}
\psi^2= \frac{\sum_{i=1}^d \mathbb{E}[\vert <\nabla_v, n_i >\vert^2]}{\sum_{i, \psi_i} \mathbb{E}[\vert <\nabla_v, n_i >\vert^2]}
\label{eq:14}
\end{equation}

where the denominator aggregation within a singular vector $n_i$ is determined by non-zero values, specifically when $\psi_i \neq 0$. The proportion outlined in Eq. (\ref{eq:14}) signifies the ratio of the projected gradient of $\nabla_v$ relative to the overall gradient, ensuring it remains distinct from the null-space or the kernel of the matrix $\hat W$. This assumption can be approximated as ${d}/{\min(d, c)}$. Following this premise, approximately ${\min(d, c)}/{d}$ of the gradient $\nabla_v$ will reside within the $\min(d, c)$-dimensional subspace. Consequently, for the purpose of norm regularization, we adjust the singular values to $\sqrt{d/\min(d, c)}$. Nevertheless, direct implementation of $d/\min(d, c)$ is computationally intensive as the square root operation involving matrices and poor implementation could potentially disrupt the training process. In order to address this concern, we draw inspiration from \cite{nie2022iteratively} and adopt an iterative algorithm for matrix square root computation through matrix multiplication. This approach ensures efficient model training, as the iterations only involve matrix multiplication.
%%%%%

\subsection{DWTNet Training} In our model, the unmasked pixels are used to recover the corresponding pixels of the input image $x_{\mathfrak m}$, while the latent vectors $ {\mathcal V}\in \mathbb R^{k\times c}$, that are responsible for feature vector $\hat f$ is interpreted from an image and use to recover the masked and unmasked pixels, where $k$ and $c$ represent the total number of latent vectors and the dimensional of feature vectors respectively. This allows the decoder learns to reconstruct image $x_{r}$ from the input $x_{\mathfrak m}$, thus we can write the loss as:

%-----------------
\begin{equation}
\ell_{DWTNet}=\underbrace{\ell_{R}(x_{\mathfrak m},x_{ r})}_1+ \beta \underbrace{\Vert \hat \nabla [\mathcal V] \ominus \hat f \Vert_2^2}_2,
\label{eq:2-1}
\end{equation}

\noindent in which $\hat \nabla$[.] is a pause-gradient operation that stops gradients from flowing into its argument, and $\beta=0.25$. The second term of Eq. (\ref{eq:2-1}) is the commitment loss \cite{van2017neural} that transfers gradient information from decoder to encoder.
%%%%%%%%%%%%%%%%%%%%%%%%%%%%
The first term indicates the reconstruction loss $\ell_{R}$(.,.), which determines the dissimilarity between the corrupted and reconstructed images. It is composed of five components, including the $\ell{1}$ between two images' pixel values ($\ell_{pixel}$), gradients ($\ell_{G}$), the adversarial loss ($\ell_{A}$), the perceptual loss ($\ell_{P}$), and the style loss ($\ell_{S}$) between the two images. We were inspired by \cite{nazeri2019edgeconnect} for the designs of the last three losses. Following is a detailed description of the losses listed above.

%-----------------
\begin{equation}
\ell_{pixel}=\mathcal M(|x_{\mathfrak m}\ominus x_{ r}|),
\label{eq:2-4}
\end{equation}
%---------------
\begin{equation}
\ell_{G}=\mathcal M(|\nabla [x_{\mathfrak m}]\ominus \nabla [x_{ r}]|),
\label{eq:2-5}
\end{equation}

\noindent in which $\mathcal M(.)$ denotes a mean-value operation and $\nabla[.]$ is the function that calculates the image gradient. The adversarial loss $\ell_{A}$ is calculated using a discriminator network $\mathcal D_{A}(.)$:
%-----------------
\begin{equation}
\ell_{A}=-\mathcal M(\text {log}[1\ominus\mathcal D_{A}(x_{r})])- \mathcal M(\text {log}[\mathcal D_{A}(x_{\mathfrak m})]),
\label{eq:2-6}
\end{equation}

\noindent $\log[.]$ represents the element-wise logarithm procedure. The network architecture of the discriminator is identical to that described in \cite{nazeri2019edgeconnect}. On the basis of the activation maps from VGG-16, the conceptual loss $\ell_{P}$ and style loss $\ell_{S}$ are computed.

%------------------
\begin{equation}
\ell_{P}=\sum_{l}^{L_P} \mathcal M(|\rho_l (x_{\mathfrak m})\ominus \rho_l (x_{ r})|),
\label{eq:2-7}
\end{equation}
\begin{equation}
\ell_{S}=\sum_{l}^{L_S}\mathcal M(|\mathcal G(\rho_l (x_{\mathfrak m}))\ominus \mathcal G(\rho_l (x_{ r}))|),
\label{eq:2-8}
\end{equation}
%---------------
\noindent in which $\rho_l(.)$ denotes different layers in VGG-16, and $\mathcal G(.)$ represents the function that returns the Gram matrix of its argument. For $\ell_{P}$ and $\ell_{S}$, $L_P$ is set to \{lrelu1-1, lrelu2-1, ..., lrelu5-1\}, and $L_S$ is set to \{lrelu2-2, lrelu3-4, lrelu4-4, lrelu5-2\} respectively. $\ell_{R}$ is therefore equal to the sum of the above four losses. %($\ell_{pixel}+\alpha_{g}\ell_{G}+\alpha_{a}\ell_{A}+\alpha_{p}\ell_{P}+\alpha_{s}\ell_{S}$). 
%
%---------------
%\begin{equation}
%\ell_{R}=\ell_{pixel}+\alpha_{g}\ell_{G}+\alpha_{a}\ell_{A}+\alpha_{p}\ell_{P}+\alpha_{s}\ell_{S}.
%\label{eq:2-9}
%\end{equation}
%---------------
%%%%%%%%%%%%%%%%%%%%%%%%%%%%%%
\begin{table}
\centering
\footnotesize{
\caption{\footnotesize{Details of our network architectures, in which} $ch$ \footnotesize{stands for base channel width. We apply LRelu (0.1), Conv $3\times3$, and Tanh at all dimensions for the output layer.}}
\label{tab:3}
\resizebox{9cm}{!}{
\begin{tabular}{l|l}     \hline\hline
\rule{0pt}{1\normalbaselineskip}{Encoder} & {Decoder}  \\  [0.2ex]  \hline
\rule{0pt}{1\normalbaselineskip}RGB image $x \in \mathbb {R}^{256\times256\times3}$ & $z \in \mathbb {R}^{8\times8\times4. ~ch}$  \\[0.1ex] \hline
\rule{0pt}{1\normalbaselineskip}Conv $3\times3$  &-                           \\[-0.2ex]
Down Resnet $128 \times 128 \times 1~ . ~ch$ & Up Resnet $8 \times 8 \times 4~ . ~ch$    \\ [-0.2ex]  
Down Resnet $64 \times 64 \times 2~ . ~ch$ & Up Resnet $16 \times 16 \times 4~ . ~ch$    \\ [-0.2ex]
Down Resnet $32 \times 32 \times 4~ . ~ch$ & DWT block $\times 4~$ + Res-FFC $\times 2~$   \\ [-0.2ex]
DWT block $\times 4~$ & Up Resnet $32 \times 32 \times 4~ . ~ch$ + Res-FFC               \\[-0.2ex]
Down Resnet $16 \times 16 \times 4~ . ~ch$ & Up Resnet $64 \times 64 \times 2~ . ~ch$    \\ [-0.2ex]
Down Resnet $8 \times 8 \times 4~ . ~ch$ & Up Resnet $128 \times 128 \times 1~ . ~ch$ + Res-FF     \\ 
 \hline
\end{tabular}}
}
  \end{table}
%%%%%%%%%%%%%%%%%%%%%
\begin{figure}[t]
\centering
\includegraphics[width=0.99\columnwidth]{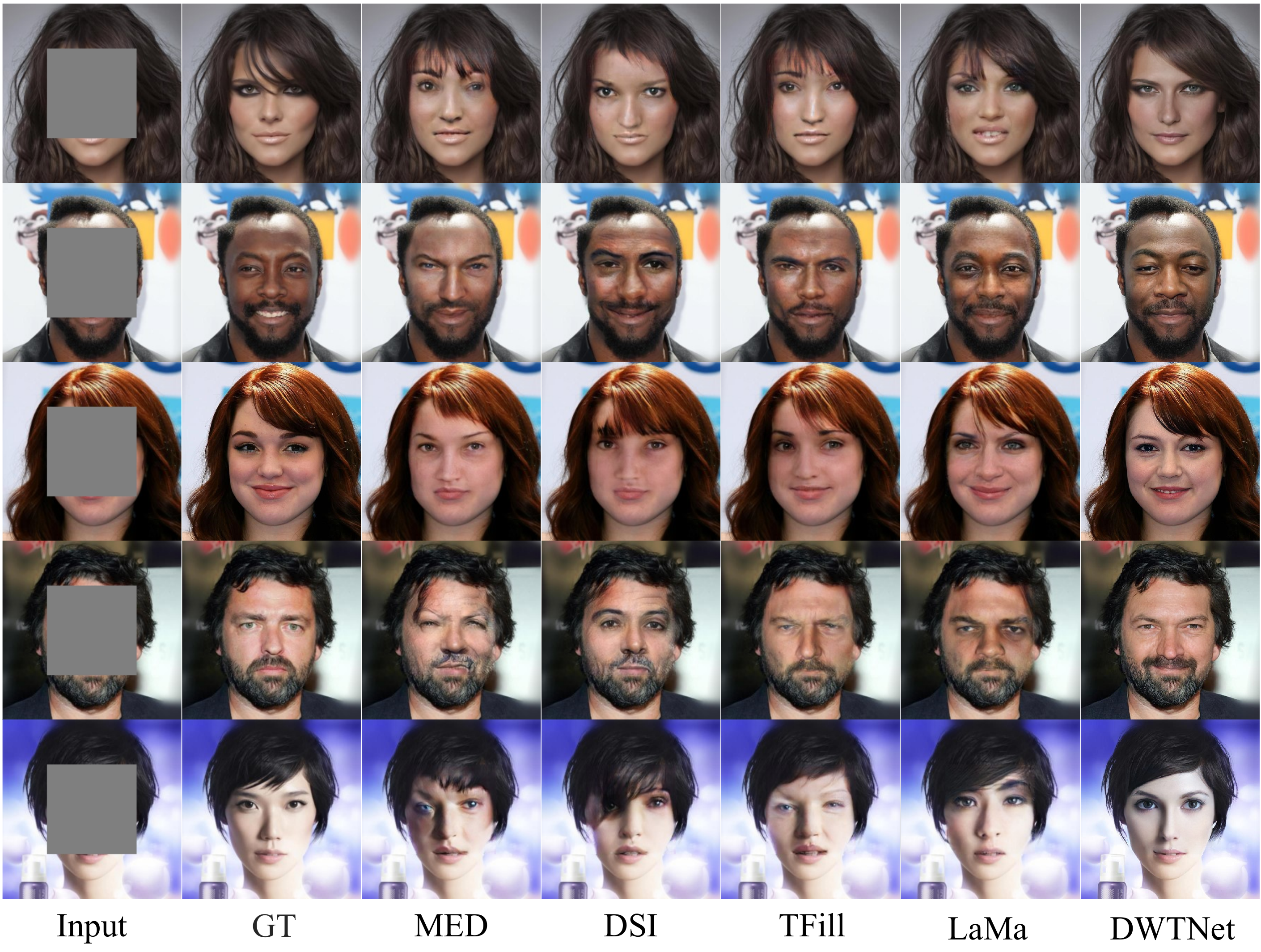} 
\caption{Qualitative comparison on the CelebA dataset. The face images produced by our model are more realistic and have more characteristic facial features as compared to other baselines.}
\label{celeba-comp}
\end{figure} 
%----------------------

\section{Experiments}
Three datasets—CelebA-HQ \cite{karras2017progressive}, Places2 \cite{zhou2017places}, and ImageNet \cite{ russakovsky2015imagenet} are used to evaluate our model, and for each dataset the standard training, testing, and validation splits are followed. For ImageNet, only 1K images from the test split are randomly selected for evaluation, the same strategy as used in ICT \cite{ wan2021high}. 
Additionally, following other image inpainting methods, we use four evaluation metrics, including U-IDS \cite{zhang2018unreasonable}, P-IDS \cite{zhao2021large}, FID \cite{heusel2017gans}, SSIM \cite{wang2004image}, and PSNR for evaluation. These metrics reflect how humans actually evaluate the quality of images.
%-----------------

\subsection{Implementation Details}
Our model is implemented in Pytorch and we use eight RTX 3080 GPU to train the model for approximately 170 hours with a batch size of 20 for 1M iterations. For the implementation, we set the loss weight to $\alpha_{g}$ = 5, $\alpha_{a} = \alpha_{p}$ = 0.1, $\alpha_{s}$ = 250. In all networks, spectral normalization is applied. Moreover, Orthogonal Initialization is used to initialize the networks, and the networks are trained at a fixed learning rate of 1e-4 and the Adam is used with $\beta_1$ = 0 and $\beta_2$ = 0.9. We also use Adam to optimize the transformer with a fixed learning rate of 3e-4. To ensure a fair comparison with earlier inpainting approaches, all images for training and testing are $256 \times 256$ in size, with regular or irregular missing regions in random locations. The details of our network architecture is shown in Table \ref{tab:3}.

%%%%%%%%%%%%%%%%%%%%%%%%%%%
\subsection{Performance Evaluation}
We compare the performance of our model with state-of-the-art inpainting models. DFv2 \cite{yu2019free}, EC \cite{nazeri2019edgeconnect}, MED \cite{liu2020rethinking}, DSI \cite{peng2021generating}, TFill \cite{ zheng2022bridging}, CoMod-GAN \cite{zhao2021large}, and LaMa \cite{suvorov2022resolution} using the provided pre-trained weights.

\begin{figure}[b]
\centering
\includegraphics[width=0.99\columnwidth]{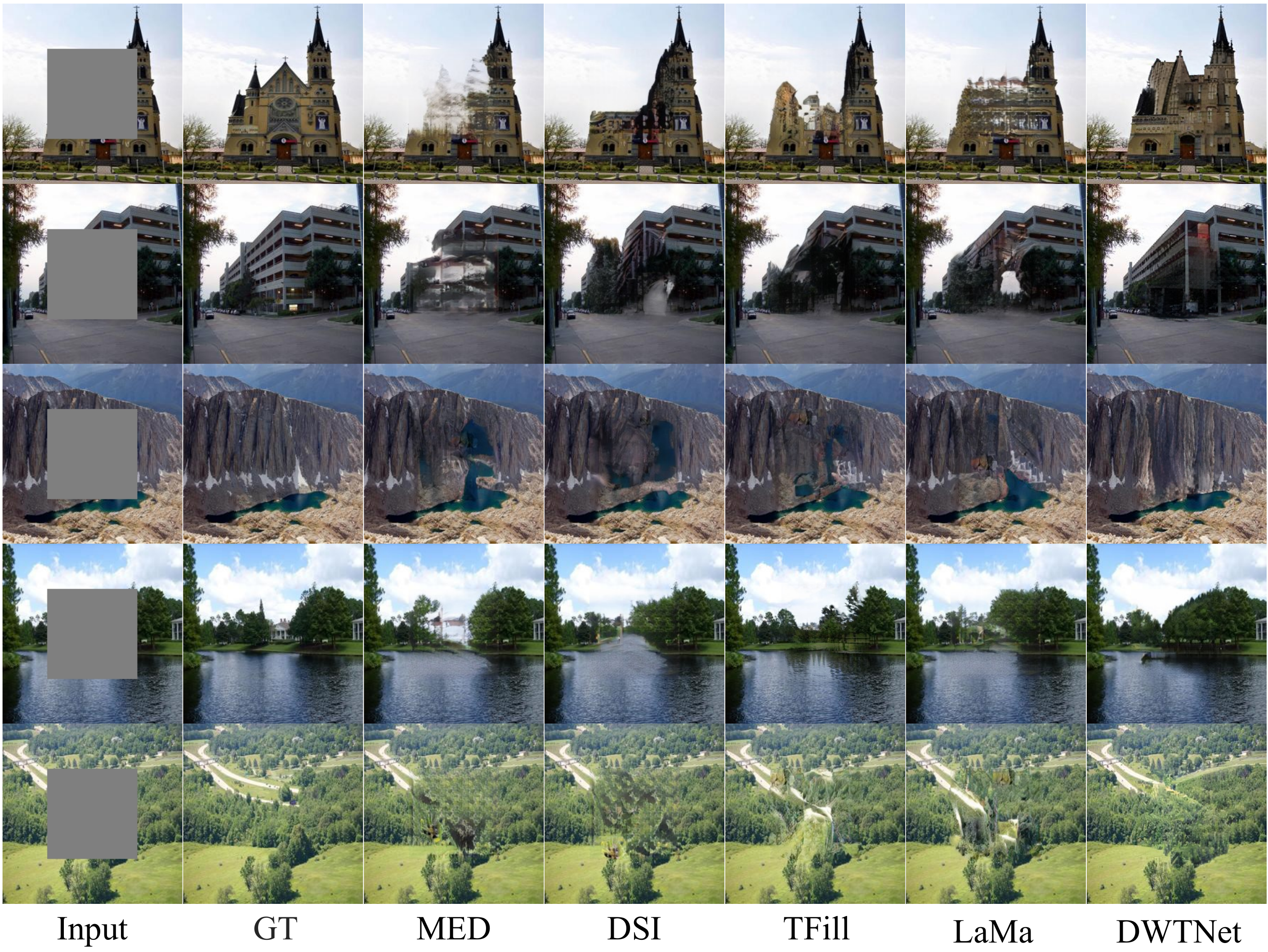} 
\caption{Qualitative comparison on the Places2. Our model is successful at reducing blur and artifacts produced by inconsistencies in structure and texture within and around missing regions.}
\label{place-comp}
\end{figure}

\begin{figure}[t]
\centering
\includegraphics[width=0.99\columnwidth]{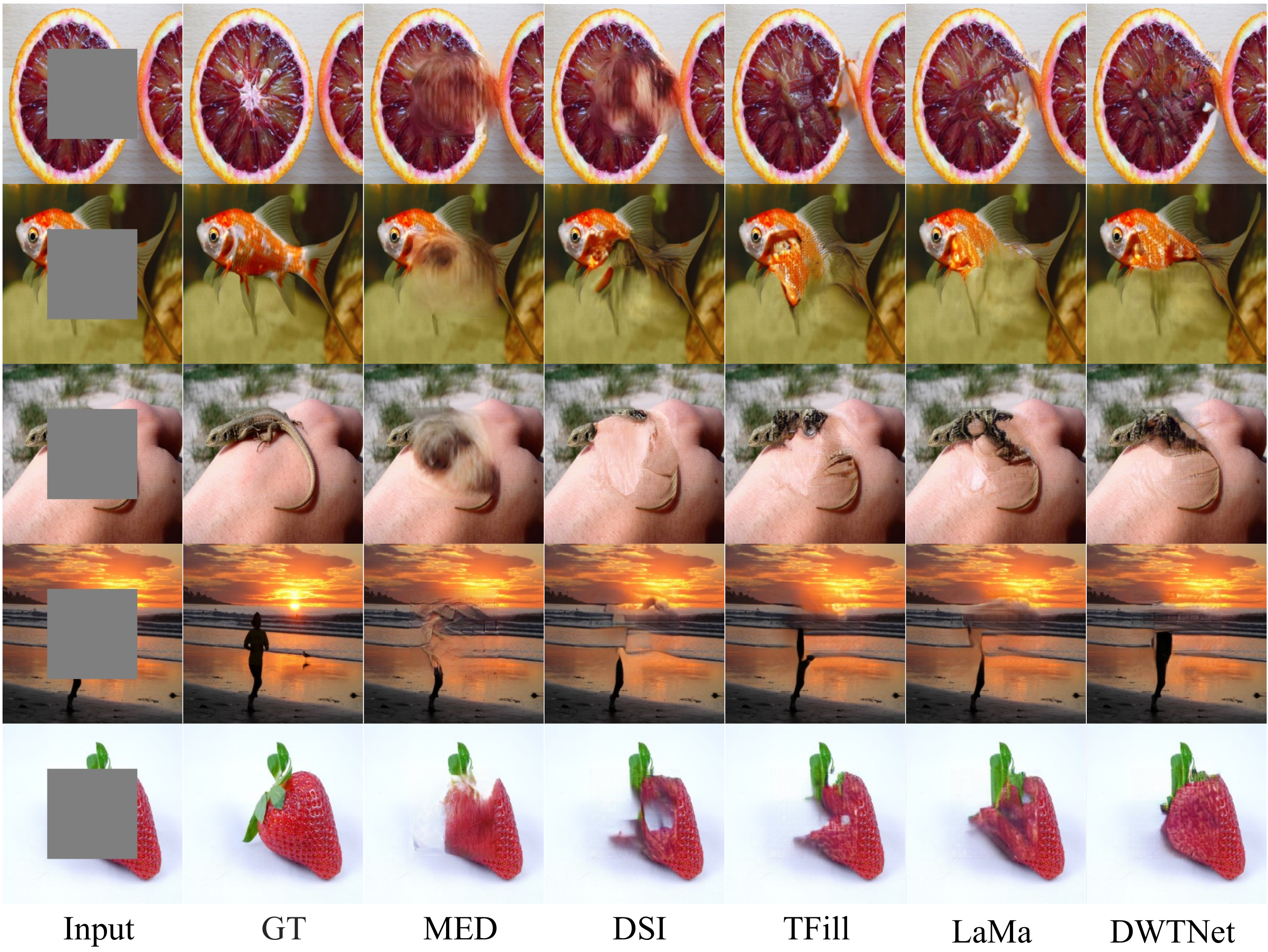} 
\caption{\footnotesize{Qualitative comparison on the ImageNet dataset. Our model outperforms existing methods in terms of retaining both structures and textures.}}
\label{imagenet-comp}
\end{figure}

\noindent{\bf Qualitative Comparisons:} Center-hole inpainting's qualitative comparison results on CelebA, Places2, and ImageNet are shown in Figs.~\ref{celeba-comp},~\ref{place-comp}, and \ref{imagenet-comp}. In comparison to other methods, MED tends to produce implausible structures, and DSI generates textures with unnatural artifacts. Moreover, in the DSI model, the contribution of structural information to texture synthesis is limited. In contrast, our method proves to be more effective than TFill and LaMa in understanding the global context and preserving realistic textures, especially when dealing with challenging images from datasets like Places2 and ImageNet. We believe that the excellence of our model can be attributed to two main factors: 1) The DWT modules that play a crucial role in capturing long-range dependencies and understanding global context in the input images. 2) The Res-FFC modules that facilitate the optimal reuse of high-frequency features, enhancing the generation of repeating textures. These components help preserve essential information from the input images, leading to the generation of more realistic and meaningful completions.
%%%%%%%%%%%%%%%%
%%%%%%%%%

%----------------------
%----------------------
%-----------------------------------
\begin{table*}
%\footnotesize{
%\renewcommand{\arraystretch}{1.1} 
\centering
\caption{\footnotesize{Quantitative comparisons on CelebA, Places2, and ImageNet datasets on center masked images. Results are based on P-IDS ($\uparrow$), U-IDS ($\uparrow$), FID ($\downarrow$), and SSIM ($\uparrow$). DWTNet w/o Res-FFC indicates our model that use two convolution layers to aggregate $\mathfrak{f}$ and $\mathfrak{f}_{\text{skip}}$. DWTNet w/o DWT indicates our model trained with the standard ViT. Among the baselines, LaMa and CoModGAN are the closest competitors to ours. However, both of them are considerably more complex than ours, while also exhibiting inferior performance. This comparison highlights our method's superior efficiency in using trainable parameters and achieving faster inference (Inf.) speeds per image. }}

\label{tab:1}
\resizebox{18cm}{!}{
\begin{tabular}{l|c c c c|c c c c|c c c c|c c}     \hline\hline
%\multicolumn{5}{|c|}{Results on training on untransformed data (0)} \\  \hline
\rule{0pt}{1\normalbaselineskip}\multirow{ 2}{*}{Method} & \multicolumn{4}{c|}{CelebA}& \multicolumn{4}{c|}{Places2}& \multicolumn{4}{c|}{ImageNet} & \multicolumn{2}{c}{Complexity} \\   \cline{2-15}   
             & \rule{0pt}{1\normalbaselineskip}P-IDS & U-IDS & FID & SSIM   & P-IDS & U-IDS & FID & SSIM & P-IDS & U-IDS & FID & SSIM& Params (M)& Inf. (s)  \\    \hline

\rule{0pt}{1\normalbaselineskip}DFv2  & 5.25& 13.46& 9.52& 0.846  & 5.18& 13.18& 17.83& 0.752 & 2.24& 7.95& 23.42& 0.685& 4 & 0.3  \\ [-0.6ex]
EC  & 6.08& 15.79& 8.16& 0.859  & 5.62& 14.45& 17.27& 0.774 & 2.68& 7.81& 23.76& 0.694& 22 & 1 \\[-0.6ex]  
MED  & 8.76& 22.17& 7.45& 0.873 & 8.33& 20.14& 13.65 & 0.792 & 4.12& 10.63& 18.84& 0.726& 18& 0.2   \\ [-0.6ex]    
DSI  & 10.84& 23.15&  7.12& 0.911 & 10.24& 21.53& 12.93& 0.829 & 4.73& 11.57& 17.25& 0.775& 40 & 7   \\ [-0.6ex]
%ICT  & 6.68 & 16.24&  5.65 & 5.92& 15.71& 9.28& 5.67& 14.21& 14.63   \\ [-0.6ex]
TFill  &10.91  &24.29 &  5.31& 0.917 & 10.31& 21.94& 9.02& 0.840 & 5.89& 16.76& 11.81& 0.791& 69& 4  \\ [-0.6ex]
CoMod-GAN  &11.26  &24.83 &  5.03& 0.921 & 10.79& 22.66& 8.70& 0.847 & 6.75& 17.42& 11.26& 0.805& 109& 5    \\ [-0.6ex]
LaMa  & 11.57& 25.12&  4.87&0.925 & {11.64}& {23.51}& 8.56& 0.851 & 6.92& 17.74& 10.61& 0.810& 51  & 3   \\                                                       \hline 
%DWTNet w/o CT * & 0.101 &  {6.34} &{0.908}& {0.130}&  {10.97} & {0.844}& {0.166}& {11.25}& {0.776} \\  
DWTNet w/o Res-FFC & 10.43 &  22.98 & 7.24& 0.894 & 9.78&  20.42 & 13.07& 0.817 & 4.38& 11.25& 17.63& 0.761& - & - \\ 
DWTNet w/o DWT & 11.02 &  24.53 & 5.12& 0.919& 10.86&  22.73 & 8.96& 0.848 & 6.64& 17.15& 11.49& 0.792& - & - \\ 
DWTNet (ours) & \bf{12.08}& \bf{26.34}&  \bf{4.51}& \bf{0.939} & \bf{12.14}& \bf{23.98}&  \bf{8.04}& \bf{0.868} & \bf{7.35}& \bf{18.26}&  \bf{10.12}& \bf{0.824}& 42& 2  \\   \hline
\end{tabular}
}
\end{table*}
%-------------------------
\begin{figure}[t]
\centering
\includegraphics[width=0.99\columnwidth]{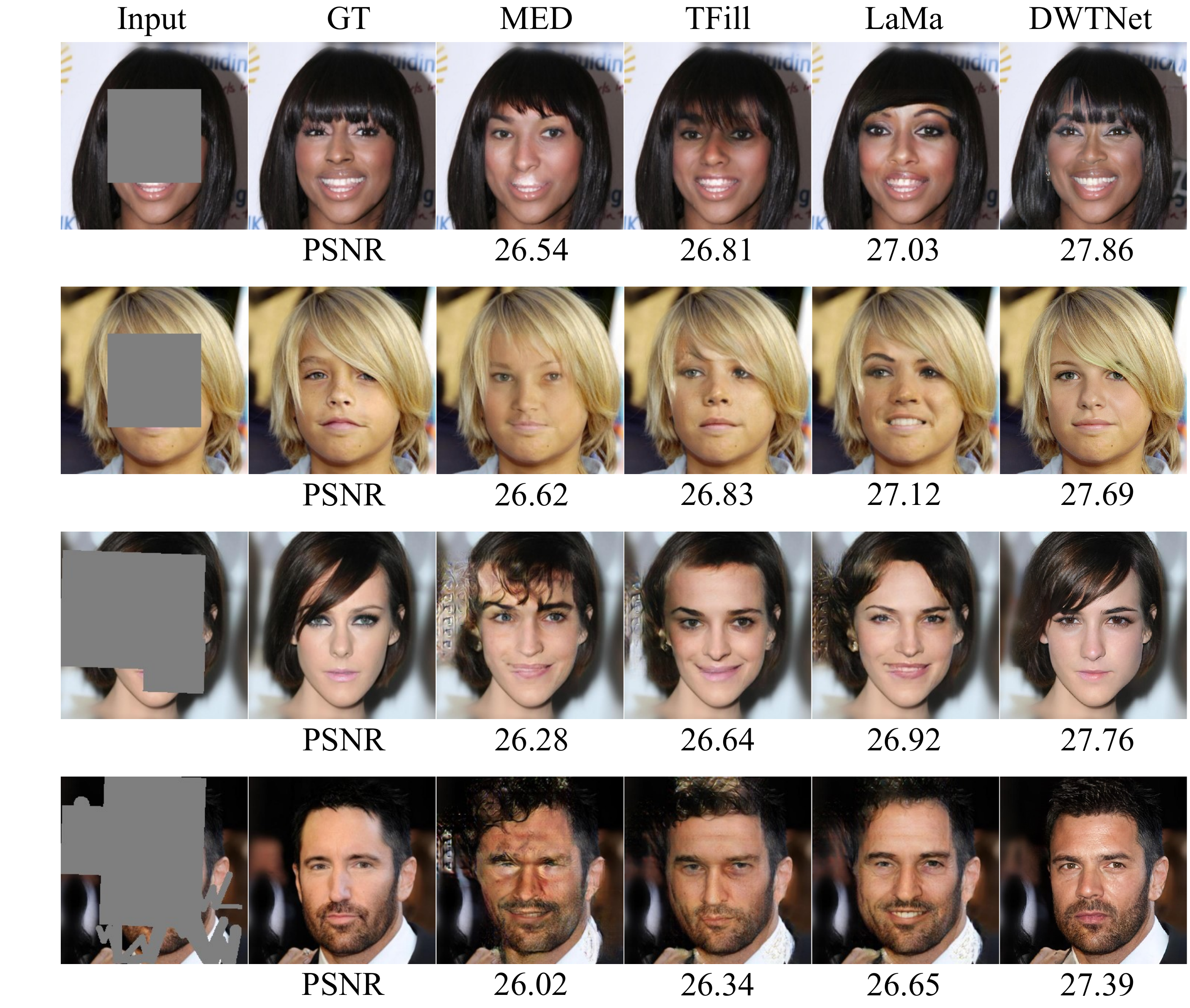} 
\caption {Visual results comparison of our model with other models on CelebA with PSNR values. A higher PSNR implies less distortion.}
\label{celeba-PSNR}
\end{figure}
%%%%%%%%%%%%%%
%%%%%%%%%%%%%%
\begin{figure*}
\centering
\includegraphics[width=1.75\columnwidth]{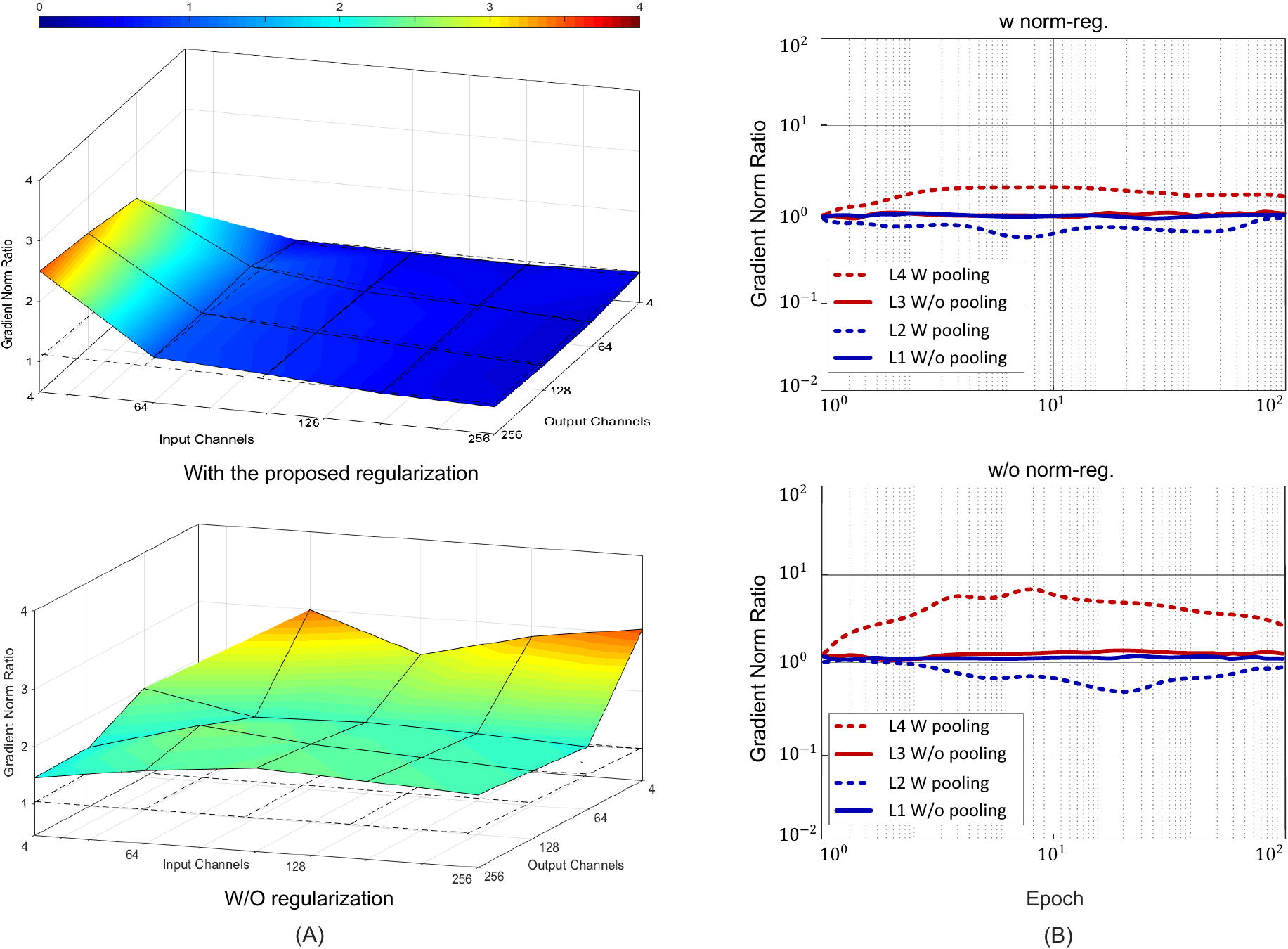} 
\caption{Proposed norm-regularization method evaluation. (A) The gradient norm ratio for different input and output channels with and without using the proposed regularization method. (B) The layers without pooling are indicated by solid lines, while the layers with pooling are indicated by dashed lines.}  
\label{fig:11}
\end{figure*}

\noindent{\bf Quantitative Comparisons:}
We compare our DWTNet with several baseline methods. For a fair comparison, we test the models on the same masks. On all three datasets, as reported in Table \ref{tab:1}, DWTNet achieves higher or equivalent results. In this experiment, all test samples with $128 \times 128$ center masks are used as the comparison's basis. 
By using DWT and Res-FFC networks, our DWTNet learns to generate texture details in the output image without any distortions or blurriness. In particular, without the DWT module, our model cannot have satisfactory performance on images with complex textures. For example, without the DWT module on Places2, our model has a FID of 8.87. On the other hand, without the Res-FFC, DWTNet struggles with texture reconstruction (FID of 17.63 on ImageNet). 
As demonstrated in Figs. \ref{place-comp} and \ref{imagenet-comp}, MED and DSI are not good at reconstructing the images with large missing regions, however, our DWTNet produces more realistic images with fewer artifacts as compared to other approaches. Fig. \ref{celeba-PSNR} shows some visual examples of how our model produces more realistic textures without adding extra artifacts. Fig. \ref{celeba-PSNR} further demonstrates that, despite having high PSNR values, TFill and LaMa are unable to produce visually plausible images while there is a large missing region, which means they cannot compete with their generative networks. DWTNet excels in capturing and generating complex periodic structures. Moreover, it achieves these capabilities with fewer trainable parameters and inference time costs compared to competitive methods.
%------------------------------------------------

\subsection{Norm Regularization}
To evaluate the impact on performance of our regularization method, the following experiment is conducted. The regularization is performed on a small network with four convolution layers (similar to encoder). In this experiment, the odd layers have $3\times3$ convolutions, and the even layers have $1\times1$ convolutions. Fig. \ref{fig:11}(A) represents the norm of gradient ratio changes for various input {\it c} and output {\it d} channels at the $20^{th}$ iteration on CelebA, with and without using our norm-regularization method. The averaged results over $5$ runs show that our model improves norm preserving capability, as it improves the ratio of gradient norms toward $1$ which resulting in better performance and faster convergence.  
%%%%%%%%%%%%%%%%%%
\begin{figure}
\centering
\includegraphics[width=0.98\columnwidth]{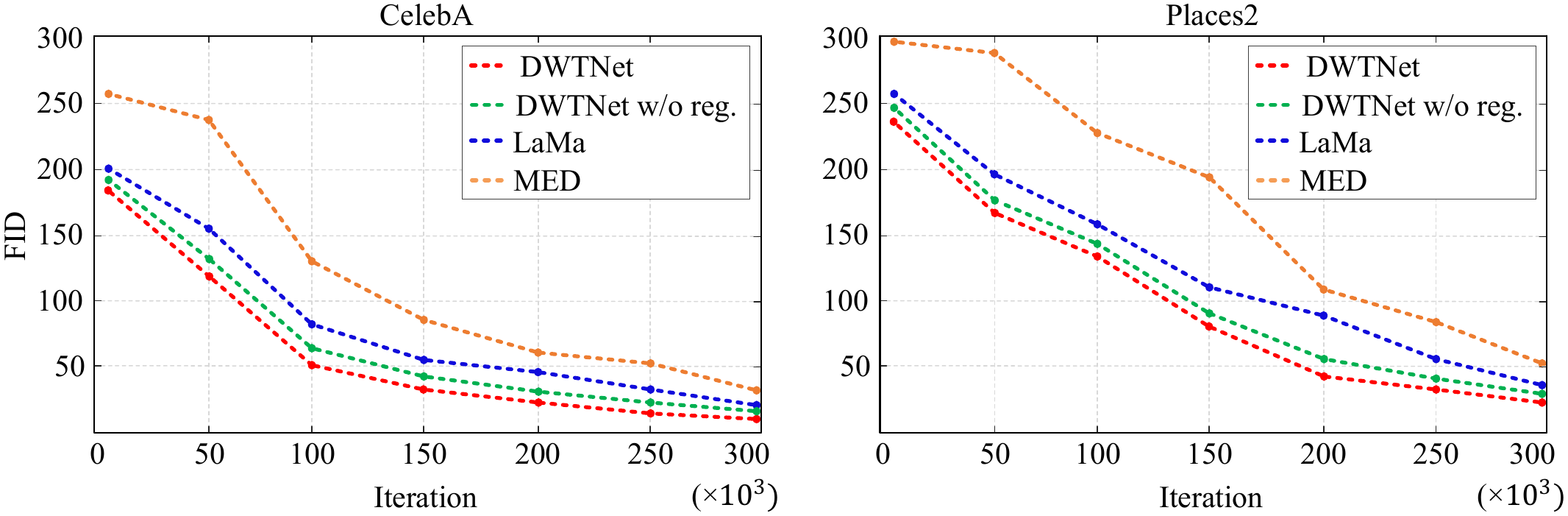} 
\caption{\footnotesize{The FID score for CelebA and Places2 generated for our model with and without norm-regularization, LaMa, and MED as a function of iterations.}}  
\label{FID}
\end{figure}
%%%%%%%%%%%%%%%%%

In addition, the FID measure in Fig. \ref{FID} demonstrates that, in comparison to the baselines, our model with norm-regularization has a faster convergence. For example, our model enhances the FID by 76 and 93 on CelebA and Places2 after 40k iterations. In Fig. \ref{fig:11}(B), the ratios of the network with/without norm-regularization are evaluated during the training. In this experiment, the encoder consists of four layers, two layers containing pooling operations that alter the dimensions and the other two without pooling. The network is trained for 100 iterations.
We emphasize the significance of initialization in constructing a norm-preserving network. Although the model without norm-regularization initially maintains norm preservation, the gradient's range gradually increases and deviates from the desired value of 1 as the parameters are updated over time. This observation underscores the importance of proper initialization and norm regularization for stable and effective training.
%This results in a stronger norm-preserving ability compare to the standard convolutions.
%------------------------------------------
%\begin{figure}
%\centering
%\includegraphics[width=0.73\columnwidth]{fig7} 
%\caption{\footnotesize{Proposed norm-regularization method evaluation. It shows the ratios of gradient norms in the first 100 epochs. The solid lines denote the without pooling layers, where the dashed lines are the layers with pooling. }}  
%\label{fig:7}
%\end{figure}  %

\subsection{Human Perceptual and Ablation Study}
We conduct a user study to obtain a more precise evaluation of subjective image quality compared to DSI, TFill, and LaMa. We randomly select 30 masked images from the CelebA and Places2 test sets. For each image, we generate two reconstructed outputs: one using DWTNet and the other using one of the baselines. Participants in the study are presented with both reconstructed images simultaneously and asked to choose the one that appears most photorealistic and visually natural.
We gather results from 27 participants and calculate the preference ratios for each approach based on the data provided in the Table. \ref{user-study}(A). For CelebA and Places2 our method have a 61.8\% and 68.4\% likelihood of being selected, respectively.
%%%%%%%%%%%%%%%%%%%%%
\begin{table*}
\centering
\caption {Ablation study. Table (A) shows the results of a study on human perception. Table (B) shows different configurations of our model on Places2. Type \enquote{A}: indicates our full architecture. Type \enquote{B}: replacing the transformer with CNNs. Type \enquote{C}: replacing our DWT blocks with the standard transformer. Type \enquote{D}: indicates the use of our model without an autoregressive decoder. Types \enquote{E} and \enquote{F} : indicate the performance of our model with Res-FFC while local and global branches are disabled, respectively.}
\includegraphics[width=1.75\columnwidth]{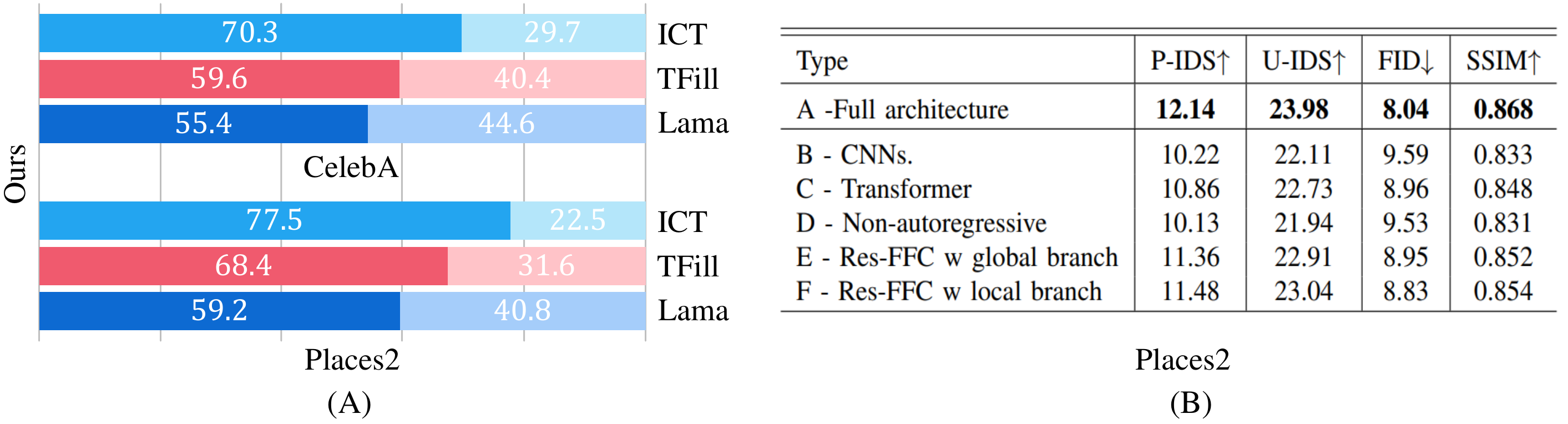} 
%The values represent the preference for the comparison pair on CelebA-HQ and Places2.}}
\label{user-study}
\end{table*}
%%%%%%%%%%%%%%%%%%%%%%%%%%%%%%%%%%%%%%%
%-----------------
We also perform ablation studies on our model to analyse which elements of our proposed architecture most significantly influence the overall performance. In order to conduct this investigation, we train the models using 100K training images from Places2. For testing, we randomly select 10K validation images. In Table \ref{user-study}(B), the quantitative comparison is reported.
%------------------------------------
\begin{figure}[t!]
\centering
\includegraphics[width=0.97\columnwidth]{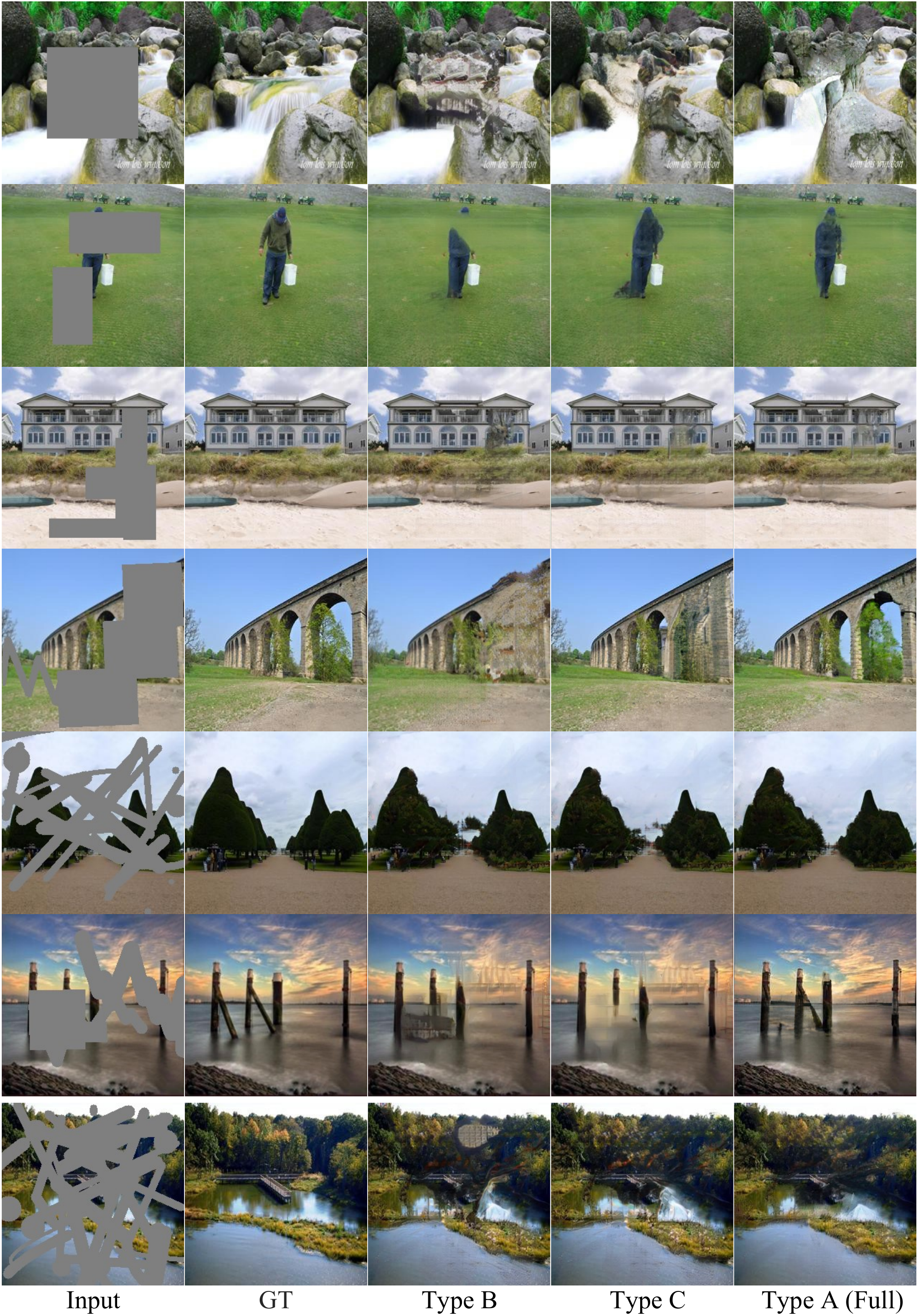} 
\caption{Ablation study examples. Type A represents our comprehensive model, whereas B and C are simplified versions that use convolutions in place of transformers and the original transformer.}
\label{placeR-comp}
\end{figure}

%\begin{table}[h]
%\centering
%\footnotesize{
%\caption{\footnotesize{Ablation study of our model components.  Type "A" denotes our full architecture. Convolutions replace transformers in "B". In "C" we replace our proposed DWT blocks with the original transformer} \cite{vaswani2017attention}. \footnotesize{In "D" we show the performance of our model without autoregressive decoder.} \footnotesize{In "E" we show the performance of our model without Res-FFC.}}
%\label{tab.4}
%\begin{tabular}{l|c|c|c}     \hline\hline
%\rule{0pt}{1\normalbaselineskip}Type & P-IDS$\uparrow$ & U-IDS$\uparrow$&  {FID}$\downarrow$  \\  [0.2ex]  \hline 
%\rule{0pt}{1\normalbaselineskip}A -Full architecture & \bf{12.14 } & \bf{23.98 } & \bf{8.04}  \\ \hline 
%\rule{0pt}{1\normalbaselineskip}B - Trans.   & 10.22  &  22.11& 9.59    \\  
%C - DWT blocks   & 11.56  &  23.62& 8.61    \\ 
%D - Non-autoregressive   & 10.13  &  21.94 & 9.53  \\  
%E - Without Res-FFC   & 11.47  &  23.25 & 8.68  \\   
%\hline
%\end{tabular}}
% \end{table}
%----------------------------

\noindent{\bf Transformer-Convolution Framework.} We investigate if or not a transformer with multi-head self-attention is effective for filling large missing regions. The inpainted images lose some quality when the transformer blocks are replaced with convolution blocks (Table \ref{user-study}(B) type \enquote{B}), as seen the performance reduction on all the metrics. Additionally, we demonstrate a few examples in Fig. \ref{placeR-comp}. In comparison with the convolution network, our model uses distance context to reconstruct visual structure, demonstrating the efficiency of long-term interactions.
\vspace{2pt}

\noindent{\bf DWT Block.} We developed a new transformer block in our framework because the standard design is prone to unstable optimization. As shown in Table \ref{user-study}(B), our full model (type \enquote{A}) outperforms type \enquote{C} with the original transformer \cite{vaswani2017attention}, improving performance by 0.83 on FID. As seen in Fig. \ref{placeR-comp}, our model generates visually more pleasing images, allowing for high-quality image completion.
\vspace{2pt}

\noindent{\bf Autoregressive Decoder.} We observed that autoregressive modeling is a key component of learning data distributions and it represents a significant component to produce high-quality images. In Table \ref{user-study}(B) \enquote{D} we demonstrate the effect of autoregressive decoder as a data synthesizer in the performance of our model.
\vspace{2pt}

\noindent{\bf Res-FFC.} In two different architectures, we evaluate the effect of this module. (1) In our architecture, we combine the encoder and decoder features without using the Res-FFC blocks (two convolution layers are used to aggregate the features). (2) Connecting the skipped features of the encoder with the Res-FFC blocks to the generator features. The quantitative comparison in Table \ref{tab:1} demonstrates the significance of Res-FFC blocks (improves the model performance on CelebA by 2.36 FID).
Moreover, to evaluate the impact on local and global branches of Res-FFC, we performed a comparison in Table \ref{user-study} \enquote{E} and \enquote{F}. In these studies we analyze our model's behavior when each branch is disabled. %providing insights into the importance of local and global information in the image completion process. 
Our ablation study confirms that both the local and global branches play crucial roles in improving our model performance.

%%%%%%%%%%%%%%%%%%%
\section{Conclusion}
In this paper, we introduced DWTNet for image completion. To enhance the quality of the reconstructed images, we proposed DWT with an autoregressive VAE to calculate the weights for image tokens and encode global dependencies. Additionally, we designed Res-FFC by integrating coarse features from the generator with the encoder's skip features, which helps the model to generalize on unseen images. Indeed, the Fourier convolutions of FFC provide a high receptive field  which result in better perceptual quality. Furthermore, to stabilize training a norm-preserving method is proposed to improve performance. Without using singular value decomposition, we introduced a simple regularization method to determine the nonzero singular values of the convolution operator. We experimentally show our model has significant potential for content generation, because of its ability to approximate soft relationships between image contents.
DWTNet achieved other baselines' performance on several benchmarks. Our architecture has been shown to be superior via extensive quantitative and qualitative comparisons. Although DWTNet achieves better results than current state-of-the-art approaches on images that have been damaged with regular and irregular masks, it still struggles when analyzing objects with different shapes or complex textures. For instance, the DWTNet outperforms competing approaches on the ImageNet dataset, however, there is still room for further improvement in the quality of the generated images. Consequently, we will work on a more advanced Transformer to fully comprehend the semantic content of an image.


%\newpage


{\footnotesize{
\bibliographystyle{IEEEtran}
\bibliography{IEEEabrv,IEEEexample}
}}
%

\end{document}